\documentclass[letterpaper]{article} 
\usepackage{aaai24}  
\usepackage{times}  
\usepackage{helvet}  
\usepackage{courier}  
\usepackage[hyphens]{url}  
\usepackage{graphicx} 
\usepackage{natbib}  
\usepackage{caption} 
\usepackage{algorithm}
\usepackage{algorithmic}

\graphicspath{{figure/}}
\usepackage{subfigure}

\usepackage{epsfig}
\usepackage{amsmath}

\usepackage{amssymb}
\usepackage{multirow}
\usepackage{cite}
\usepackage[table]{xcolor}
\usepackage{bbding}
\usepackage{enumerate}
\usepackage{pifont}
\usepackage{booktabs}
\usepackage{wasysym}
\usepackage{url}
\usepackage{float}

\usepackage{newfloat}
\usepackage{listings}
\DeclareCaptionStyle{ruled}{labelfont=normalfont,labelsep=colon,strut=off} 
\floatstyle{ruled}
\newfloat{listing}{tb}{lst}{}
\floatname{listing}{Listing}
\title{IBAFormer: Intra-batch Attention Transformer for Domain Generalized Semantic Segmentation}

\author{Qiyu Sun$^{1}$, Huilin Chen$^{1}$, Meng Zheng$^{2}$, Ziyan Wu$^{3}$, Michael Felsberg$^{4}$, Yang Tang$^{1}$\\
$^{1}$ East China University of Science and Technology, $^{2}$ Rensselaer Polytechnic Institute, $^{3}$ United Imaging Intelligence, $^{4}$ Linköping University \\
{\tt\small y20190063@mail.ecust.edu.cn, 2454007639@qq.com, mengzhengrpi@gmail.com, ziyan.wu@uii-ai.com, michael.felsberg@liu.se, yangtang@ecust.edu.cn}
}

\usepackage{bibentry}

\begin{document}

\maketitle

\begin{abstract}

Domain generalized semantic segmentation (DGSS) is a critical yet challenging task, where the model is trained only on source data without access to any target data. 
Despite the proposal of numerous DGSS strategies, 
the generalization capability remains limited in CNN architectures.
Though some Transformer-based segmentation models show promising performance, they primarily focus on capturing intra-sample attentive relationships, disregarding inter-sample correlations which can potentially benefit DGSS. 
To this end, we enhance the attention modules in Transformer networks for improving DGSS by incorporating information from other independent samples in the same batch, enriching contextual information, and diversifying the training data for each attention block.
Specifically, we propose two alternative intra-batch attention mechanisms, namely mean-based intra-batch attention (MIBA) and element-wise intra-batch attention (EIBA), to capture correlations between different samples, enhancing feature representation and generalization capabilities. 
Building upon intra-batch attention, we introduce IBAFormer, which integrates self-attention modules with the proposed intra-batch attention for DGSS. 
Extensive experiments demonstrate that IBAFormer achieves SOTA performance in DGSS, and ablation studies further confirm the effectiveness of each introduced component.

\end{abstract}

\section{Introduction}

To date, utilizing synthetic data for training deep networks has gained increasing attention in various computer vision applications like autonomous driving, healthcare, and surveillance~\cite{richter2016playing,ros2016synthia,STRAPS2020BMVC}. 
It is particularly crucial for applications like semantic segmentation where acquiring large-scale real data and pixel-level annotations are highly time- and labor-demanding.
However, segmentation models trained on synthetic data may generalize poorly to real-world scenarios due to domain gaps~\cite{torralba2011unbiased}. 
To address this, domain adaptation (DA) methods~\cite{hoffman2018cycada,ma2021coarse} and domain generalization (DG) approaches~\cite{yue2019domain,choi2021robustnet} have been proposed.
DA methods require access to a specific target domain for joint training, which may not be feasible in practical applications, while DG-based models are trained solely on source domain data, making them more relevant in practice. Our work focuses on the most challenging task, \textit{i.e.}, synthetic-to-real, domain-generalized semantic segmentation (DGSS)~\cite{peng2022semantic}. 

Existing DGSS approaches are predominantly based on convolutional neural networks (CNNs)~\cite{yue2019domain,choi2021robustnet}.
Since recent advancements have demonstrated the superiority of Vision Transformers compared to CNNs in out-of-distribution generalization~\cite{paul2022vision,wenzel2022assaying}, 
it is natural to ask if Transformers can improve the performance of DGSS.
Note that existing Transformer-based semantic segmentation models are assembled with multiple self-attention layers~\cite{xie2021segformer,hoyer2022daformer}, and they ignore potential correlations across different samples. 
We argue that such correlations introduce variability to the training process, which brings better generalizability of the model.
Furthermore, by leveraging cross-sample information, the model gains access to more diversified contextual and structural cues not present in individual images, facilitating the training process to be aware of the overall structure of the scene~\cite{seidenschwarz2021learning}.







The primary challenges in utilizing cross-sample information are selecting appropriate samples for augmentation and effectively integrating cross-sample information for more generalized model learning. 
While video semantic segmentation~\cite{Li_2018_CVPR} has leveraged consecutive frames and improved performance using temporal correlations, the assumption of having access to consecutive sequences is restrictive and may not apply to all scenarios. 
Additionally, semantic segmentation is a single-image prediction task, making it challenging to utilize different samples without temporal correlation.
To address these challenges, we randomly sample images without temporal correlation and propose a novel attention mechanism called intra-batch attention to fuse cross-sample information.
Specifically, we introduce two alternative intra-batch attention: mean-based intra-batch attention (MIBA) and element-wise intra-batch attention (EIBA).
{In contrast to self-attention modules, which analyze the associations of tokens within a single sample, EIBA computes the associations with different samples in the batch and focuses on the most relevant one.
In an alternative method, MIBA computes associations between an image and the aggregate mean value of other samples within a batch in a more computationally efficient way.}
Both MIBA and EIBA effectively leverage information and encode contextual cues across samples in the same batch.
Based on intra-batch attention, we introduce IBAFormer, composed of several intra-batch attention and self-attention modules, to extract intra-batch relationships in both low-level and high-level features. 
Experiments show that our IBAFormer leads to a significant improvement in DGSS.
The contributions of our work can be summarized as follows:

\begin{itemize}
   
    \item [1)] 
    We introduce two intra-batch attention mechanisms, namely MIBA and EIBA, to integrate intra-batch contextual information.

    \item [2)] Building on intra-batch attention, we propose IBAFormer, which merges intra-batch features at multiple levels.

    
    \item [3)] Extensive experiments clearly demonstrate that IBAFormer exhibits SOTA performance in DGSS and ablation studies validate the effectiveness of each component introduced.

\end{itemize}


\section{Related Work}

We review the literature on Vision Transformers and highlight their potential advantages in robustness for visual tasks. Subsequently, we discuss relevant work in DGSS, along with Transformer-based segmentation approaches.

\textbf{Vision Transformers}
ViT~\cite{dosovitskiy2021an} emerges as a preferred architecture for various visual tasks due to its ability to achieve SOTA performance in numerous visual tasks~\cite{nicola2018end,dosovitskiy2021an,liu2021swin,wang2021pyramid,wang2022crossformer}, as well as its robustness to distribution shifts.
The core of ViT is self-attention, and extensive research has focused on designing efficient self-attention mechanisms. 
Swin Transformer~\cite{liu2021swin} introduces shifted windows for hierarchical feature maps, which facilitates connections between consecutive self-attention layers.
Tokens-to-Token ViT~\cite{yuan2021tokenstotoken} aggregates neighboring tokens into one token iteratively, which encodes information about local structure.
CvT~\cite{wu2021cvt} blends CNN and Transformer with convolution for multi-level feature capturing.

Since self-attention primarily focuses on the relationship between local patches within a single image, researchers have begun exploring more effective attention modules to capture broader relationships.
BViT~\cite{li2023bvit} introduces broad attention to exploit attention relationships across different layers, and CrossViT~\cite{chen2021crossvit} employs a dual-branch architecture to extract multi-scale features from small and large patches using cross-attention.
Xu \textit{et al.}\cite{xu2022gmflow,xu2022unifying} use cross-attention mechanisms to capture cross-view interactions in matching tasks and improve the quality of extracted features.
These methods showcase the effectiveness of multi-layer/patch/view feature extraction. 
Inspired by them, we explore relationships across multiple samples, integrating information from independent samples to enhance the generalization capabilities of attention-based mechanisms.

\textbf{Domain generalized semantic segmentation} Recently, synthetic-to-real semantic segmentation has attracted a lot of attention due to its prospective application in the real-world scenarios. 
In this line of work, the segmentation network is trained on synthetic domains and then tested on unseen real target domains.
Unlike DA approaches exploiting target domain data during training, DG methods utilize only the source domain and aim to generalize to multiple unseen target domains.
Existing DG methods can be divided into normalization and whitening, and domain randomization.
Normalization/whitening methods~\cite{pan2018two,choi2021robustnet,xu2022dirl,peng2022semantic} erase domain-specific style information and learn style-invariant features.
Randomization-based methods explore unpredictable styles of target domains and generate diverse image styles for training from the source domain.
Several works perform data augmentation in image space~\cite{yue2019domain,huang2021fsdr,peng2021global,zhong2022adversarial} and some explore data manipulation in feature space~\cite{kim2021wedge,tjio2022adversarial,lee2022wildnet,zhao2022style,wu2022siamdoge,huang2023styleprojection}.
While numerous DGSS strategies are proposed, there is limited research exploring the effectiveness of Transformer-based models.

\textbf{Transformer-based segmentation} 
Due to the robustness demonstrated by Transformers compared to CNNs, Transformer-based backbones for segmentation tasks have garnered widespread attention.
Segformer~\cite{xie2021segformer} presents a powerful and efficient semantic segmentation framework that combines Transformers with lightweight multilayer perception (MLP) decoders.
DAformer~\cite{hoyer2022daformer} highlights the potential of Transformers for unsupervised domain adaptation semantic segmentation.
Mask2Former~\cite{cheng2022masked} introduces an architecture that handles different segmentation tasks by extracting localized features through constraining cross-attention within predicted mask regions.
More relatively, SHADE and HGformer adopt Transformer-based backbones for DGSS. 
SHADE~\cite{zhao2022style} generates diverse training samples by selecting basis styles from the source distribution and extending their approach to a Transformer-based model\cite{zhao2022styleextend}. 
HGformer~\cite{ding2023hgformer} introduces an explicit grouping mechanism, combining mask classification results at different scales to achieve robust semantic segmentation.
In our approach, we also utilize a Transformer-based backbone for DGSS. However, unlike methods~\cite{zhao2022styleextend,ding2023hgformer} that use off-the-shelf backbones, we make modifications to the self-attention mechanism and propose the intra-batch attention.


\section{Methodology}
\begin{figure*}[t]
	\centering
	\includegraphics[width=0.9\textwidth]{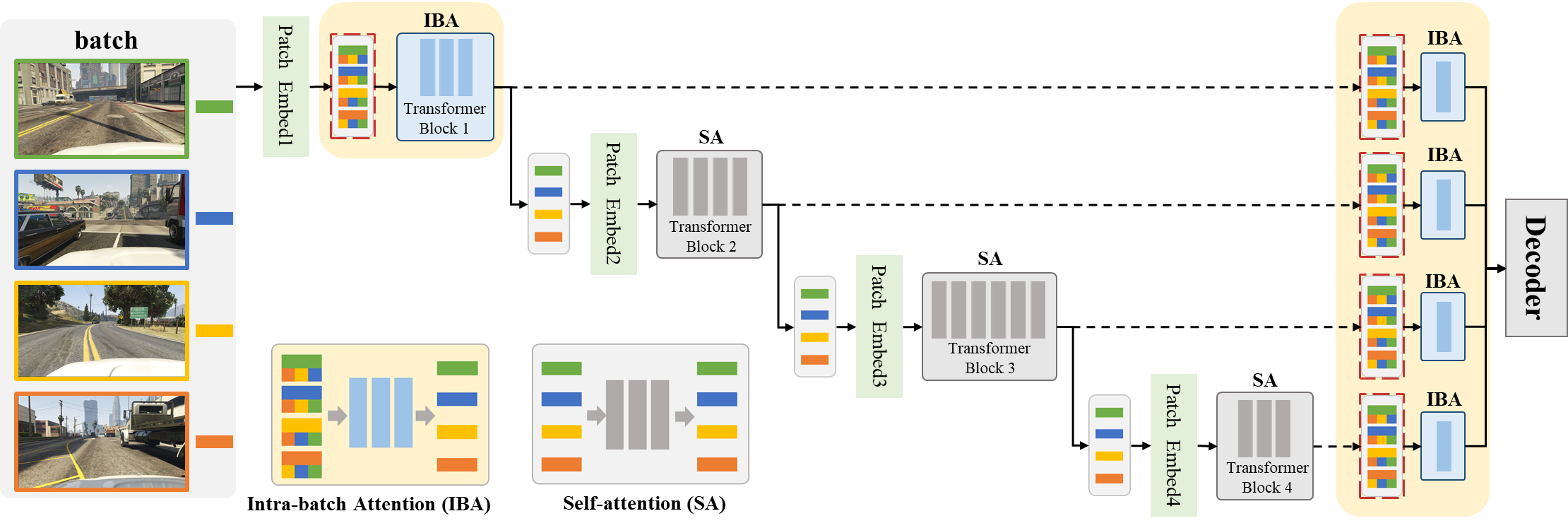}
	\caption{The training framework of IBAFormer. Batches are randomly sampled without temporal correlation. To integrate intra-batch information, we add our novel intra-batch attention to SegFormer~\cite{xie2021segformer} in two parts (yellow frames): Replacing Transformer block 1 in SegFormer with intra-batch attention modules, facilitating low-level feature fusion across the batch; and four intra-batch attention modules are added to fuse hierarchical features before the decoder. We use intra-batch information during training (red dotted frame), optionally during inference. For the specific operations for information fusion, please refer to Figure~\ref{fig:self_cross_arch}.}
	\label{fig:framework}
\end{figure*}
Self-attention mechanisms traditionally focus solely on calculating self-affinities for individual images, neglecting potential correlations across different samples.
In this work, we delve into inter-sample correlation learning to enhance feature representation of Transformer networks for DGSS.
In contrast to video processing where consecutive image frames are highly relevant and temporally correlated, capturing inter-sample correlations is not straightforward for DGSS tasks, where data samples are independent. 
To overcome this, we introduce intra-batch attention mechanisms designed to establish connections between different images in a batch.
In the subsequent sections, we provide a comprehensive overview of the intra-batch attention mechanism and introduce our IBAFormer which incorporates intra-batch attention.

\subsection{Overview}
Our intra-batch attention Transformer (IBAFormer) builds on SegFormer~\cite{xie2021segformer} backbone, a Transformer-based model tailored for semantic segmentation.
The main difference of our proposed IBAFormer to SegFormer is that it utilizes intra-batch attention modules for multiple-level feature fusion, resulting in better domain generalizability for DGSS. 
The network architecture of IBAFormer is illustrated in Figure~\ref{fig:framework}.
Given a batch of randomly sampled images $I_{1, 2, ..., B}$ (abbreviated as $I$), where $B$ represents the batch size, we leverage $I$ as input and conduct information fusion with intra-batch attention modules.
We further increase the diversity of the samples by applying Random Image Color Augmentation (RICA)~\cite{sun2023augment}.
Subsequently, we pass the augmented images through  the backbone to extract features, denoted as ${F^{l}}$, where $l$ represents the $L$th layer output features. 
In the following, we introduce  two distinct intra-batch attention modules, \textit{i.e.}, MIBA and EIBA, and illustrate in further details of how they integrate the information contained in different images for improved DGSS. 



\subsection{Intra-batch Attention}


To blend additional samples for each image during training, we introduce MIBA and EIBA, which are detailed below.
A visual comparison between self-attention, MIBA, and EIBA is presented in Figure~\ref{fig:self_cross_arch}.

\subsubsection{Mean-based Intra-batch Attention}





For each sample $F_{i}$ in a given batch $F$, MIBA aims to establish intra-batch image relevance by identifying the corresponding \textit{auxiliary reference sample} $\hat{F}_{i}$. 
In brief, $\hat{F}_{i}$ is generated by calculating the mean of the rest samples within the batch.
Then, we calculate the relationship between $F_{i}$ and $\hat{F_{i}}$, which strengthens the global perception capability of the Transformer.
  
\begin{figure*}[t]
	\centering

	\subfigure[Self-attention]{
		\includegraphics[width=0.18\textwidth]{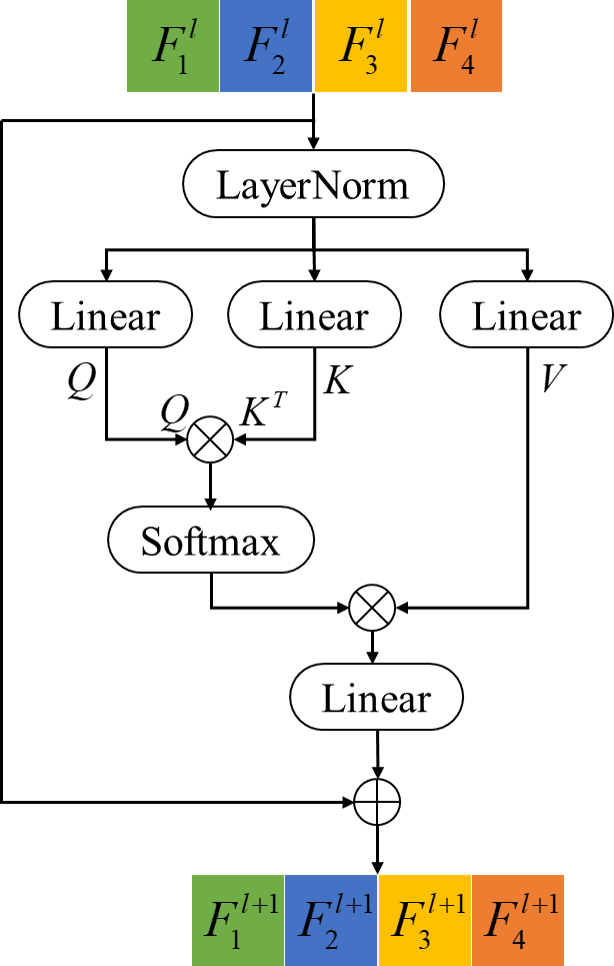}
            \label{subfig:self-attention}
	} \hspace{5mm}
  	\subfigure[MIBA]{
		\includegraphics[width=0.27\textwidth]{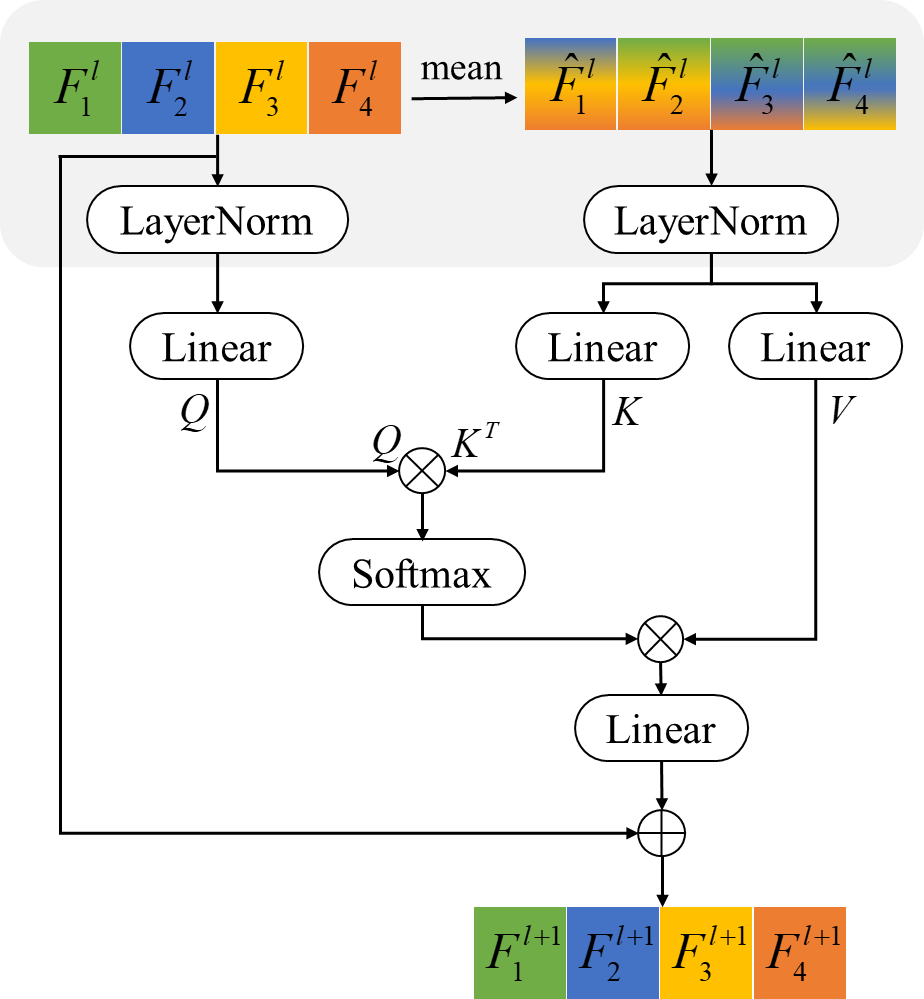}
            \label{subfig:MCBA}
	} \hspace{5mm}
   	\subfigure[EIBA]{
		\includegraphics[width=0.28\textwidth]{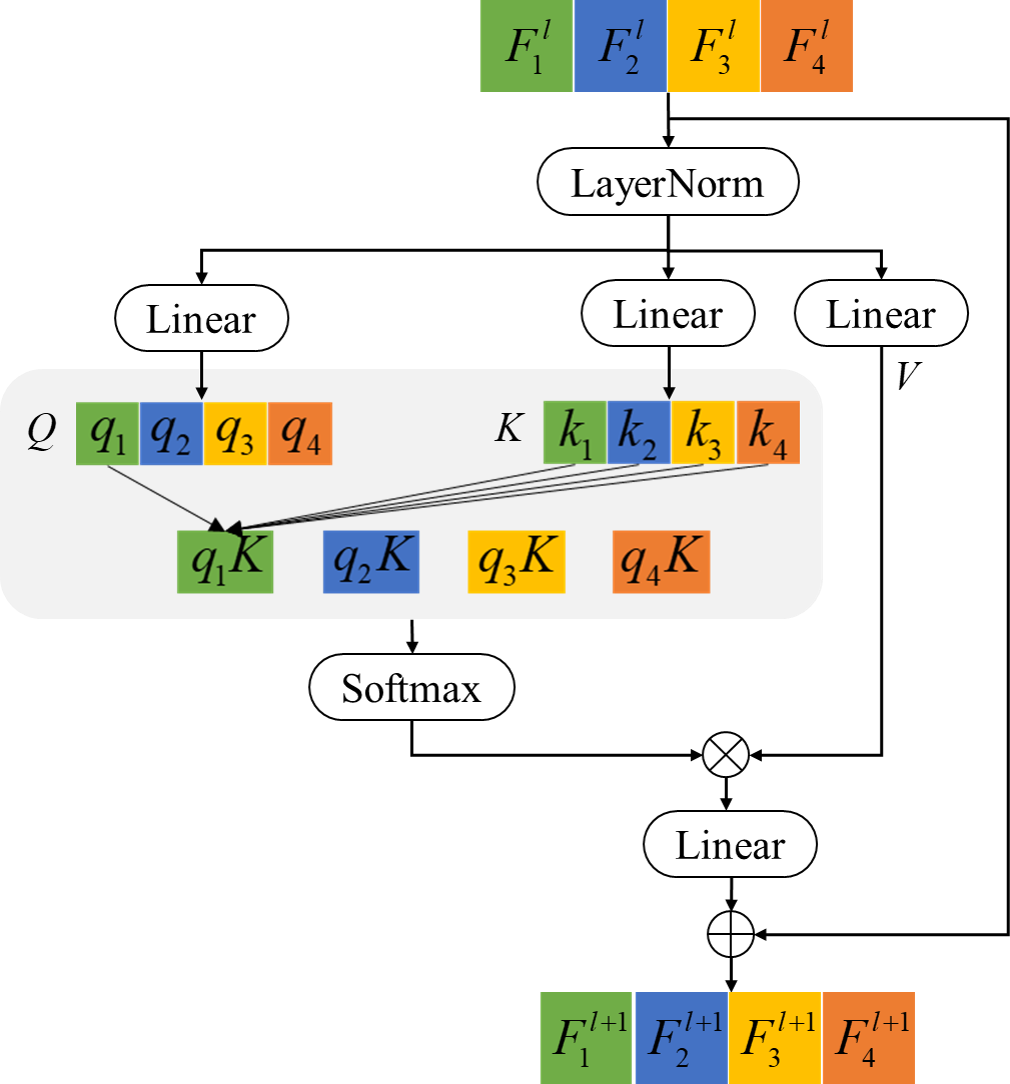}
            \label{subfig:ECBA}
	}

	\caption{The illustration of intra-batch attention modules (b) and (c), and the self-attention module (a).}
	\label{fig:self_cross_arch}
\end{figure*}

For instance,  consider the $L$th layer output features ${F}^{l}$. The generation process of the corresponding auxiliary reference batch ($\hat{F}^{l}$ for ${F}^{l}$) is illustrated in Figure~\ref{subfig:MCBA}.
The main idea is, for each feature $\hat{F}^{l}_{i}$, we calculate the mean value of other samples $F_j^{l}$ in the same batch, and use it as the reference batch $\hat{F}^{l}_{i}$:
\begin{equation}
\label{equ:mean_operation}
\begin{aligned}
\hat{F}^{l}_{i} = \frac{\sum_{j=1}^{B}{F}^{l}_{j}-{F}^{l}_{i}}{B-1}\;.
\end{aligned}
\end{equation}
By performing these operations for all samples in batch $F^{l}$, we obtain $\hat{F}^{l}$. 
Finally, $F^{l}$ and $\hat{F}^{l}$ are used as inputs to the MIBA mechanism. 
We present a pseudo-code implementation of our proposed algorithm in Algorithm~\ref{alg:MCBA}.

The MIBA follows a similar structure to self-attention:


\begin{subequations}
\label{equ:mcba1}
\begin{align}
    Q = {\rm Linear}(F)\;, \\
    K, V = {\rm Linear}(\hat{F})\;,
\end{align}
\end{subequations}


\begin{equation}
\label{equ:mcba3}
\begin{aligned}
{\rm MIBA}(Q, K, V) = {\rm Softmax}(\frac{QK^{\rm T}}{\sqrt{d_\mathrm{head}}})V\;,
\end{aligned}
\end{equation}
where $d_\mathrm{head}$ represents the number of heads in multi-head attention.
The key distinction between self-attention and MIBA lies in the calculation of $Q$, $K$, and $V$.  
In self-attention, they are all calculated through $F$. 
Conversely, in MIBA, $Q$ is calculated from $F$, whereas $K$ and $V$ are computed from $\hat{F}$. 
This allows us to capture the relationships between each image and the other images in the batch. 
By incorporating information from both ${F}$ and $\hat{F}$, we can effectively model the interactions and dependencies among all the images within the batch.
Then, we can obtain $F^{l+1}$:

\begin{equation}
\label{equ:mcba_f_l+1}
\begin{aligned}
F^{l+1} = {\rm Linear}\big({\rm MIBA}(Q, K, V)\big)+F^{l}\;.
\end{aligned}
\end{equation}

\begin{algorithm}[t]
\caption{Mean-based Intra-batch Attention (MIBA)}
\label{alg:MCBA}
\textbf{Input}: A batch of features $F^{l}$ \
\begin{algorithmic}[1] 
\STATE Calculate the reference batch $\hat{F}^{l}$ for $F^{l}$ with Eq. (\ref{equ:mean_operation})
\STATE Calculate the attention map MIBA between $F^{l}$ and $\hat{F}^{l}$ using Eq. (\ref{equ:mcba1}) - (\ref{equ:mcba3})
\STATE \textbf{return} $F^{l+1}$ in the next layer with Eq. (\ref{equ:mcba_f_l+1})
\end{algorithmic}
\end{algorithm}

\begin{algorithm}[tb]
\caption{Element-wise Intra-batch Attention (EIBA)}
\label{alg:ECBA}
\textbf{Input}: A batch of features $F^{l}$\
\begin{algorithmic}[1] 
\STATE Calculate $Q$, $K$, and $V$ using Eq. (\ref{equ:ecba1})
\FOR{each sample $q_{i}$ in $Q$}
\STATE Calculate $R(q_i, K)$  with Eq. (\ref{equ:ecba_r}) 
\ENDFOR
\STATE Calculate the attention map EIBA using Eq. (\ref{equ:ecba2})
\STATE \textbf{return} $F^{l+1}$ in the next layer with Eq. (\ref{equ:ecba_f_l+1}) 
\end{algorithmic}
\end{algorithm}

\subsubsection{Element-wise Intra-batch Attention}
As illustrated in Figure~\ref{subfig:ECBA}, in contrast to MIBA, which directly calculates the mean of the features in a batch, $Q$, $K$, and $V$ in EIBA are all calculated through $F$ like in self-attention:

\begin{equation}
\label{equ:ecba1}
\begin{aligned}
Q, K, V = {\rm Linear}(F)\;.
\end{aligned}
\end{equation}

For each sample $q_i \in Q$, EIBA computes the relationship $R(q_i, K)$ between $q_i$ and each element $k_j \in K$, and aggregates them through summation:

\begin{equation}
\label{equ:ecba_r}
\begin{aligned}
R(q_i, K)= \sum_{j=1}^{B}q_{i}k_{j}^{\rm T}\;.
\end{aligned}
\end{equation}
We can obtain $R(Q, K)$ for the entire batch by applying Eq.~(\ref{equ:ecba_r}) iteratively for each $q_i$, and ${\rm EIBA}(Q, K, V)$ can be formulated as:

\begin{equation}
\label{equ:ecba2}
\begin{aligned}
{\rm EIBA}(Q, K, V)  
= {\rm Softmax}(\frac{R(Q, K)}{\sqrt{B}})V \;.
\end{aligned}
\end{equation}
For the input $F^{l}$, we can obtain $F^{l+1}$:
\begin{equation}
\label{equ:ecba_f_l+1}
\begin{aligned}
F^{l+1} = {\rm Linear}\big({\rm EIBA}(Q, K, V)\big)+F^{l}\;.
\end{aligned}
\end{equation}
We outline our algorithm in Algorithm~\ref{alg:ECBA} for brevity. 

\subsection{IBAFormer}
Since previous studies have demonstrated the superior performance of Transformer-based models in out-of-distribution scenarios~\cite{hoyer2022daformer,xie2021segformer}, we adopt the main network architecture from SegFormer~\cite{xie2021segformer} as our baseline.
SegFormer captures dependencies within a single image with self-attention modules, while IBAFormer utilizes intra-batch attention blocks to promote the integration of different samples by fusing them, which allows the model to capture the global structure of different images.
Please note that the samples are randomly sampled during the IBAFormer training process without temporal constraints, allowing for practical feasibility in various scenarios.
The key distinctions between IBAFormer and SegFormer are twofold:
(1) replacing the self-attention mechanism in Transformer block 1 with intra-batch attention for low-level feature fusion, 
and (2) introducing multiple intra-batch attention modules before the hierarchical features are fed into the decoder for multi-level feature fusion. 
Both MIBA and EIBA can be utilized for intra-batch attention fusion.

\subsubsection{Low-level feature fusion}
The original hierarchical Transformer encoder in SegFormer consists of four Transformer blocks that utilize self-attention modules. 
In our IBAFormer, we replace the self-attention module in the first Transformer block with intra-batch attention to facilitate low-level feature fusion. 
By leveraging this intra-batch attention mechanism, IBAFormer surpasses the performance of SegFormer. 
After conducting experiments, we found that replacing the self-attention module in the first block is the best choice for our IBAFormer model in consideration of accuracy and extra pre-training. 


\subsubsection{Multi-level feature fusion}
\label{sec:self_vs_intra}


In addition to fusing cross-sample information at the low level, IBAFormer extends its capability to incorporate multi-level information from different samples. 
By introducing intra-batch attention modules before the hierarchical feature maps reach the decoder, the network benefits from utilizing multi-layer features for cross-sample perception while preserving the pre-trained model parameters.

\begin{figure}[t]
	\centering
	\includegraphics[width=0.45\textwidth]{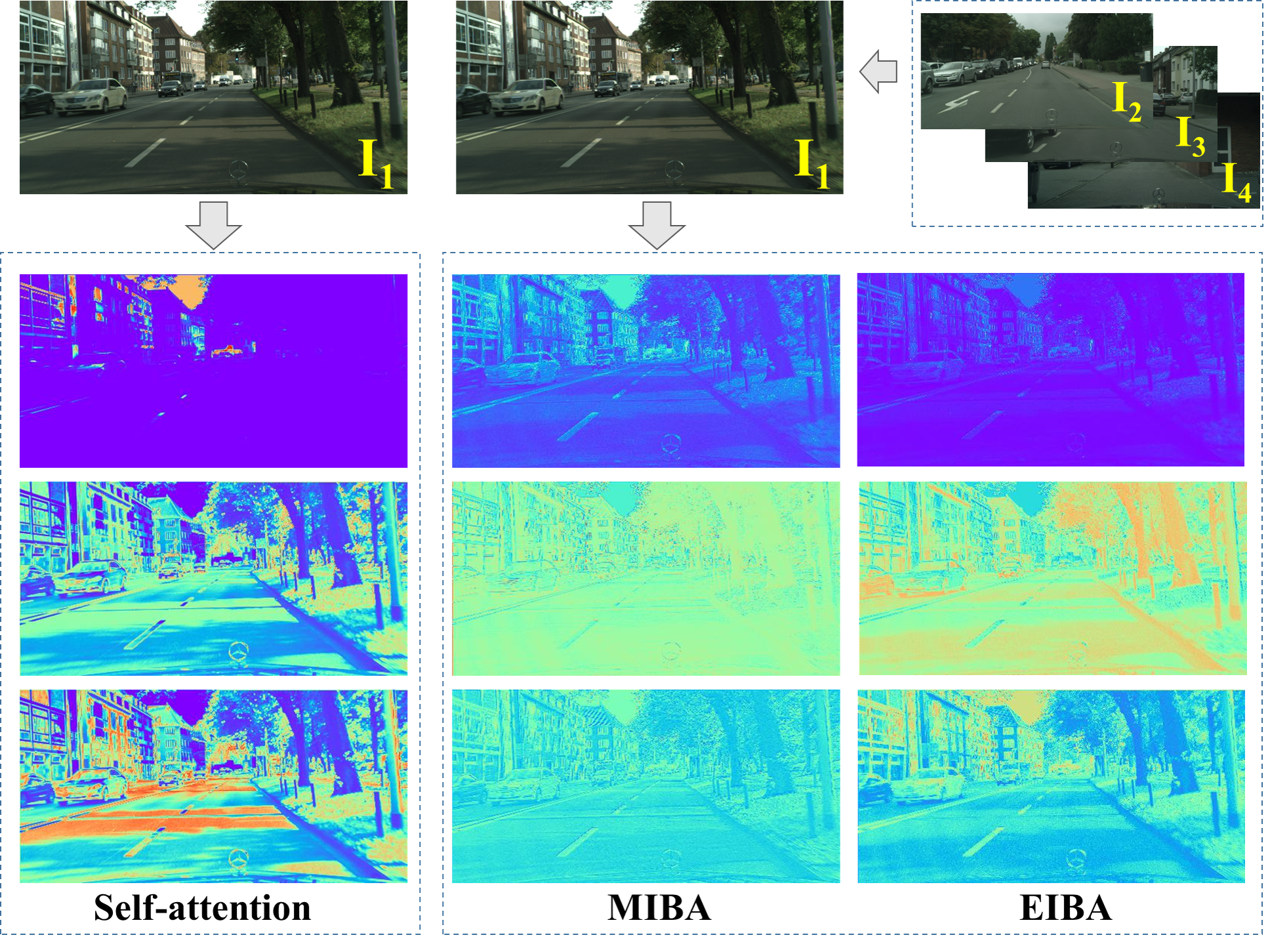}
	\caption{The illustration of the attention maps generated by self-attention, MIBA, and EIBA. 
 We choose three representative attention channels from the three attention modules, and the results indicate that self-attention tends to emphasize local regions, whereas intra-batch attention prioritizes the entire scene. 
 More attention maps can be found in the supplementary materials.
}
	\label{fig:atte_features}
\end{figure}

Through the integration of intra-batch attention, our IBAFormer model significantly improves its capability to capture semantic context across different images.
We illustrate the attention maps generated by
self-attention, MIBA, and EIBA in Figure~\ref{fig:atte_features}. 
We select three representative attention channels from the output of the first attention module in Transformer Block 1 in Figure~\ref{fig:framework}, and visualize them for comparison.
Figure~\ref{fig:atte_features} illustrates that intra-batch attention modules attend to the whole scene structure, while self-attention centers on more localized regions.
Due to its local focus, self-attention exhibits sensitivity to noise and domain bias in training data, and this tendency is detrimental to generalization.
On the contrary, intra-batch attention prioritizes scene structure and disregards local noise, aiming to capture shared patterns among multiple samples, thereby mitigating noise bias in individual images.
 




\section{Experiments}

\subsection{Experimental Setup}

\subsubsection{Datasets}
Two synthetic datasets---GTAV (G)~\cite{richter2016playing} and SYNTHIA-RAND-CITYSCAPES (S)~\cite{ros2016synthia} are utilized as source domains, and three real-world datasets---Cityscapes (C)~\cite{cordts2016cityscapes}, BDDS (B)~\cite{yu2020bdd100k}, and Mapillary (M)~\cite{neuhold2017mapillary} are used as target domains. 
In the setting of DGSS, we train the model using only the pixel-wise semantic labels in synthetic datasets, without access to images or labels in real datasets.
\textbf{Synthetic datasets:} 
GTAV contains 24,966 images extracted from Grand Theft Auto V, which is split into 12,403 training,  6,382 validation, and 6,181 testing images.
SYNTHIA has 9,400 images, and is split into 6,580 and 2,820 images for training and validation respectively~\cite{choi2021robustnet}. 
\textbf{Real datasets:} 
Cityscapes, BDDS, and Mapillary have 500, 1,000, and 2,000 images for validation, respectively.

\subsubsection{Implementation Details}
Our implementation follows the mmsegmentation framework\footnote{\url{https://github.com/open-mmlab/mmsegmentation}}. 
The MiT-B5~\cite{xie2021segformer} is used as the encoder in IBAFormer to extract a feature pyramid with $C$ = [64, 128, 320, 512].
To be consistent with ~\cite{xie2021segformer,hoyer2022daformer}, we train IBAFormer with AdamW~\cite{loshchilov2017decoupled}.
The initial learning rates of the backbone and classifier are $1\cdot 10^{-5}$ and $1 \cdot 10^{-4}$, respectively, with a weight decay of 0.01, linear learning rate warmup with 1.5k, and linear decay afterward.
We also add the rare class sampling in DAFormer~\cite{hoyer2022daformer} to SegFormer as the baseline.
The network is trained on a batch of 768 $\times$ 768 random crops for 48k iterations.
Our experiments run on 2 NVIDIA A40 GPUs with a batch of 4 for each GPU. On average, each training takes 20 hours.
Following~\cite{choi2021robustnet}, several augmentations are also applied in our experiments, such as random scale, random crop, color jitter, and random horizontal flip.

As is well known, pre-training on ImageNet significantly enhances the performance of Transformer-based models; however, the training process is time-consuming~\cite{dosovitskiy2021an}.
To circumvent the resource-intensive pre-training process, we adopt an efficient approach: initializing IBAFormer with pre-trained weights from SegFormer's ImageNet pre-training for self-attention modules and training the intra-batch modules from scratch. 
This strategy strikes a balance between the benefits of pre-training and avoiding additional pre-training for IBAFormer on ImageNet.
We evaluate the segmentation model using the mean intersection over union (mIoU).
For the models trained with GTAV, we use the 19 categories overlapping with Cityscapes for evaluation. 
When trained on SYNTHIA, we use the overlapped 16 categories for evaluation.

\subsection{Main Results}

We evaluate models in the settings of synthetic-to-real generalization and compare our method to SOTA DGSS methods quantitatively.

\subsubsection{Single-source domain synthetic-to-real generalization}
The models are independently trained on synthetic GTAV or SYNTHIA dataset and evaluated on three previously unseen real datasets: Cityscapes, BDDS, and Mapillary. 
The generalization performance of two variants of IBAFormer is shown in Table \ref{tab:gtav} (trained on GTAV) and Table \ref{tab:synthia} (trained on SYNTHIA),  respectively.
To prove the effectiveness of our proposed IBAFormer, we perform comparisons with a variety of methods that use different backbones, such as ResNet-50, ResNet-101, and MiT-B5.
Generally, our approach demonstrates superior performance over previous CNN-based methods by a substantial margin. Furthermore, it exhibits a noticeable advantage over competitive Transformer-based segmentation models with equivalent backbones.
Since our method leverages cross-sample information during training, a natural question arises: do we still need cross-sample information during inference?
To address this, we report the performance of IBAFormer with and without using intra-batch information during inference.

As demonstrated in Table~\ref{tab:gtav}, IBAFormer (EIBA) utilizing cross-sample information during inference leads to better performance on average. 
IBAFormer achieves average mIoU scores of 54.34\% and 54.79\% for MIBA-based and EIBA-based modules on three real-world target datasets, which surpasses the previous best method (SHADE) by 2.19\% and 2.64\%, respectively.
Experiment results show that, with or without utilizing cross-sample information in testing time, our model achieves SOTA performance.
This suggests that IBAFormer can be utilized in entirely new environments without any samples for augmentation, while at the same time providing great flexibility to enjoy the benefit of any available samples. Depending on the applications, it can either be used in single sample input mode, or in batch input mode to leverage the extra performance boost, when more unlabeled samples are accessible. 


The performance of IBAFormer, when trained on SYNTHIA, is also notable, as evidenced in Table~\ref{tab:synthia}. Specifically, IBAFormer equipped with the EIBA module exhibits the most robust generalization performance after training on SYNTHIA, surpassing SegFormer by {3.94}\% on CityScapes, {4.39}\% on BDD100K, and {3.94}\% on Mapillary, respectively. 
{From Table~\ref{tab:gtav} and \ref{tab:synthia}, although IBAFormer (MIBA) slightly trails behind IBAFormer (EIBA) on average, it still demonstrates impressive performance in DGSS and is implemented in a more computationally efficient way.
Since images in \textit{Mapillary} have greater variations in aspect ratios and field of view, and that MIBA prioritizes global structure, MIBA excels in this dataset in comparison to EIBA.}
These results collectively emphasize the enhanced generalizability brought about by both our MIBA and EIBA modules. In sum, our IBAFormer achieves new SOTA performance in domain-generalized semantic segmentation.





\begin{table}[t]
\caption{\textbf{Synthetic-to-real generalization results.} 
The Transformer-based methods~\cite{tang2020selfnorm,zhong2022adversarial,zhao2022styleextend} results in Table~\ref{tab:gtav} are referenced from the SHADE~\cite{zhao2022styleextend}.  
We integrate rare class sampling from DAFormer~\cite{hoyer2022daformer} into SegFormer~\cite{xie2021segformer} and implement results, using the same hyper-parameters as IBAFormer for comparisons ($\dagger$).
The remaining results are sourced from their respective original papers. 
* denotes that our IBAFormer uses intra-batch information during inference.
The reported values represent mIoU (\%).
}

\label{tab:main_result}
\centering
\subtable[G$\rightarrow$ C, B, M]{
\resizebox{\linewidth}{!}{
\begin{tabular}{c|c|ccc|c}
		\toprule[1pt]
		 Method&backbone & {G$\rightarrow$ C }  &{G$\rightarrow$ B} &  {G$\rightarrow$ M} &{Average} \\

         \toprule[1pt]
          WildNet~\cite{lee2022wildnet} & R50 &  44.62 & 38.42 & 46.09 & 43.04\\
          SHADE~\cite{zhao2022style}  & R50 &  44.65& 39.28 & 43.34 & 42.42\\
         \hline 
         WildNet~\cite{lee2022wildnet} & R101 &  45.79 & 41.73 & 47.08 & 44.87\\
          SHADE~\cite{zhao2022style}  & R101 &  46.66& 43.66 & 45.50 & 45.27\\
         \hline
         SegFormer$\dagger$~\cite{xie2021segformer}& MiT-B5 & 49.71& 47.49& 54.95 & 50.72\\
         CrossNorm~\cite{tang2020selfnorm}& MiT-B5 & 46.41& 44.69 & 50.21 & 47.10\\ 
         AdvStyle~\cite{zhong2022adversarial}  & MiT-B5 & 46.56& 45.10 & 48.35 & 46.67\\
         SHADE~\cite{zhao2022styleextend}  & MiT-B5 &  53.27& 48.19 & 54.99 & 52.15\\         
         \hline
         \textbf{IBAFormer (MIBA)}  & MiT-B5 & 55.42& 48.35 & 57.88 & 53.88 \\    
         \textbf{IBAFormer (MIBA)*}  & MiT-B5 & 55.29 & 49.16 & \textbf{58.58} & 54.34 \\ 
         \textbf{IBAFormer (EIBA)}  & MiT-B5 & 55.99& \textbf{49.78} & 58.08 & 54.62 \\
         \textbf{IBAFormer (EIBA)*}  & MiT-B5 & \textbf{56.34}& 49.76 & 58.26 & \textbf{54.79}\\
        \toprule[1pt]
\end{tabular}
}
\label{tab:gtav}
}
\subtable[S$\rightarrow$ C, B, M]{
\resizebox{\linewidth}{!}{
\begin{tabular}{c|c|ccc|c}
		\toprule[1pt]
		 Method&backbone & {S$\rightarrow$ C }  &{S$\rightarrow$ B} &  {S$\rightarrow$ M} &{Average} \\

         \toprule[1pt]
         DRPC~\cite{yue2019domain}& R50 &35.65&31.53&32.74&33.31\\
          SAN-SAW~\cite{peng2022semantic} & R50 &   {38.92}&{35.24} &{34.52}&{36.23}\\
         \hline 
          FSDR~\cite{huang2021fsdr}& R101 &40.80&{37.40}&{39.60}&{39.27}\\
         SAN-SAW~\cite{peng2022semantic}& R101 &{40.87}&35.98&{37.26}&38.04\\
         \hline
        SegFormer$\dagger$~\cite{xie2021segformer}& MiT-B5 & 46.98 &40.27 &46.64 & 44.63 \\
        \hline
        \textbf{IBAFormer (MIBA)}  & MiT-B5 & 48.36 & 42.67 & 49.16 & 46.73 \\
         \textbf{IBAFormer (MIBA)*}  & MiT-B5 & 48.17 & 43.05 & {50.12} & 47.11 \\   
         \textbf{IBAFormer (EIBA)}  & MiT-B5 & {50.02} & {44.27} & {50.04} & {48.11}  \\
         \textbf{IBAFormer (EIBA)*}  & MiT-B5 & \textbf{50.92} & \textbf{44.66} & \textbf{50.58} & \textbf{48.72} \\         
        \toprule[1pt]
\end{tabular}
}

\label{tab:synthia}
}
\end{table}

\begin{figure}[t]
 
 \centering
\begin{minipage}{0.045\linewidth}
 
 
\centerline{\includegraphics[width=\textwidth]{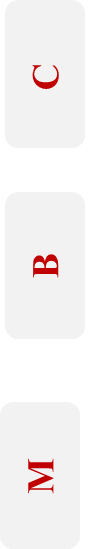}}
 
\centerline{ }
 
\end{minipage}
\begin{minipage}{0.17\linewidth}
 
 
\centerline{\includegraphics[width=\textwidth]{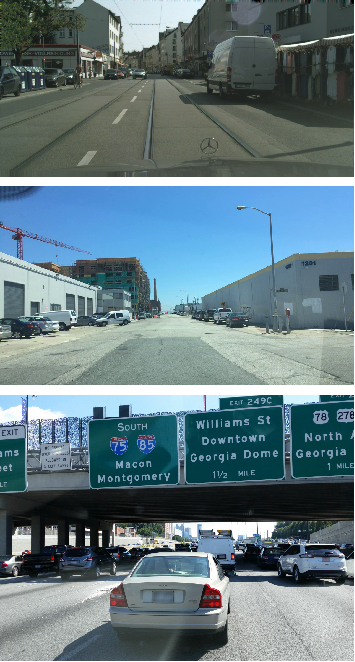}}
 \vspace{-3pt}
\centerline{\tiny{Images}}
  
\end{minipage}
\begin{minipage}{0.17\linewidth}
 
 
\centerline{\includegraphics[width=\textwidth]{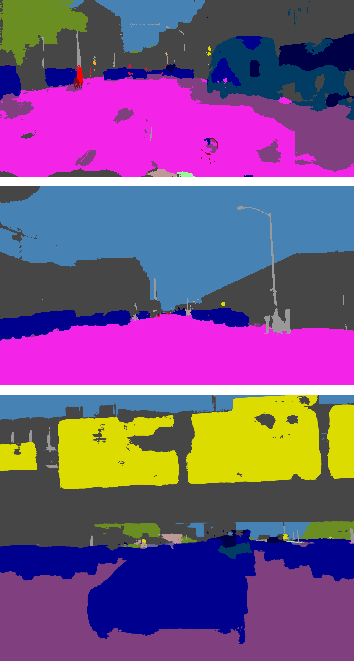}}
  \vspace{-3pt}
\centerline{\tiny{SegFormer}}
 
\end{minipage}
\begin{minipage}{0.17\linewidth}
 
 
\centerline{\includegraphics[width=\textwidth]{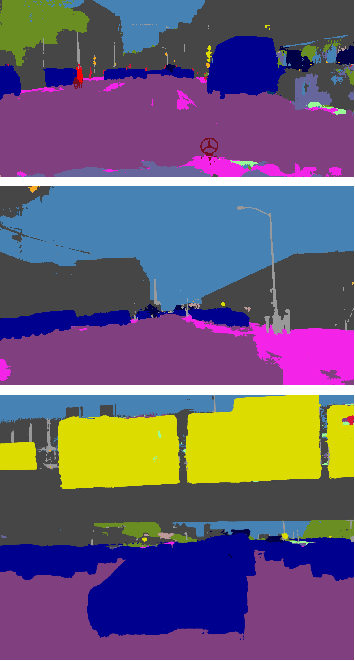}}
 \vspace{-3pt}
\centerline{\tiny{{IBAFormer(M)}}}
 
\end{minipage}
\begin{minipage}{0.17\linewidth}
 
 
\centerline{\includegraphics[width=\textwidth]{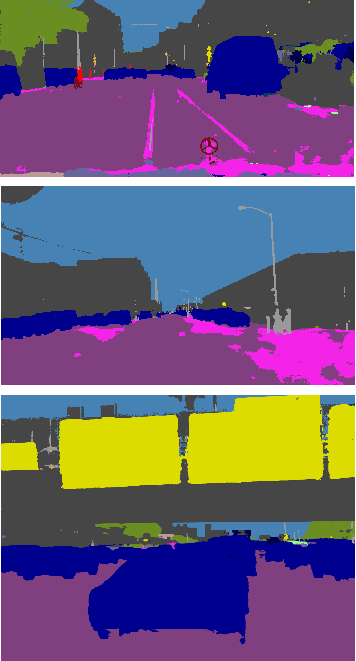}}
 \vspace{-3pt}
\centerline{\tiny{{IBAFormer(E)}}}
 
\end{minipage}
\begin{minipage}{0.17\linewidth}
 
 
\centerline{\includegraphics[width=\textwidth]{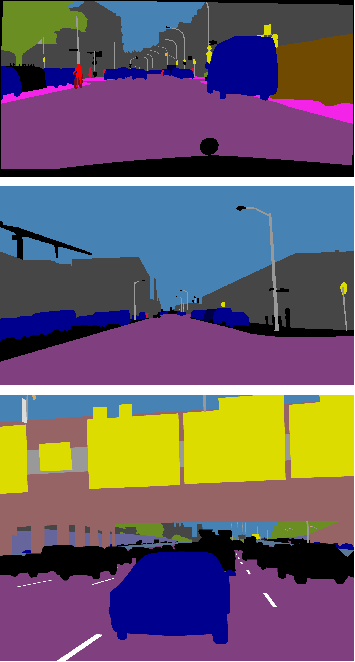}}
  \vspace{-3pt}
\centerline{\tiny{GT}}
 
\end{minipage}
\caption{Comparative visualization of SegFormer and IBAFormer (G $\rightarrow$ C, B, M).
Additional qualitative comparisons on three real datasets can be found in the supplementary materials.
}

\label{visual_comparison}
\end{figure}

We exhibit qualitative segmentation results on three target domains in Figure~\ref{visual_comparison}. As depicted in the figure, our approach effectively discriminates not only between two frequently perplexed classes, namely, \textit{sidewalk} and \textit{road}, but also other entities, such as  \textit{car}, and \textit{traffic sign}. 

Additionally, our supplementary materials provide multi-source domain generalization results, as well as the comparison of model complexity and inference time.

\begin{table}[t]
\centering
	\caption{\textbf{Ablation studies on the proposed intra-batch attention modules} (G$\rightarrow$ C, B, M).  
 The results are reported with mIoU (\%).}
	\label{tab:ablation_sum}
 \resizebox{\linewidth}{!}{
	\begin{tabular}{l|ccc|c}
		\toprule[1pt]
		{Methods}& {G$\rightarrow$ C }  &{G$\rightarrow$ B} &  {G$\rightarrow$ M} &{Average} \\

         \toprule[1pt]
         Baseline &  49.71& 47.49 & 54.95 & 50.72\\
         \hline                 
        IBAFormer (MIBA)  & 51.67 & 48.65 & 56.12  & 52.15 \\
        \hline                 
        IBAFormer (EIBA) & 53.24 & 48.09 & 55.19 & 52.17 \\	
        \hline  
        Baseline + RICA&   52.11& 48.75 & 56.76 & 52.54\\
        \hline 
        IBAFormer (MIBA) + RICA &55.29 & 49.16 & \textbf{58.58} & 54.34\\ 
         \hline                 
        IBAFormer (EIBA) + RICA  & \textbf{56.34}& \textbf{49.76} & 58.26 & \textbf{54.79} \\				     
   
        \toprule[1pt]
  
	\end{tabular}
 }
\end{table}

\subsection{Ablation Studies}
Extensive ablation experiments are conducted using the MiT-B5 encoder from GTAV to Cityscapes,
BDD100K and Mapillary on the setting of DGSS, aiming to evaluate the effectiveness of each proposed component. 
We report the results using intra-batch information during inference in all ablation studies.

\subsubsection{IBAFormer with MIBA and EIBA} 
Table~\ref{tab:ablation_sum} shows the impact of the proposed MIBA and EIBA mechanisms on DGSS. 
``Baseline" represents the SegFormer integrated with rare class sampling~\cite{hoyer2022daformer} using MiT-B5~\cite{xie2021segformer}.  
Although the baseline already performs well in DGSS, it tends to overfit to the source domain and its performance degrades on unseen domains. 
Our IBAFormer, equipped with either the MIBA or EIBA modules, achieves remarkable performance in DGSS, and the incorporation of RICA~\cite{sun2023augment} further improves the generalization capability of IBAFormer. 
As illustrated in Figure~\ref{fig:atte_features}, intra-batch attention modules attend to the whole scene structure, while self-attention centers on more localized regions.
Due to its local focus, self-attention exhibits sensitivity to noise in training data and domain bias, and this tendency is detrimental to generalization.
On the contrary, intra-batch attention prioritizes scene structure and disregards local noise, aiming to capture shared patterns among multiple samples, thereby mitigating noise bias in individual images.
These results highlight the importance of leveraging information from other samples during training, an aspect that has often been ignored in previous work.


\subsubsection{Architecture of IBAFormer} 
We experimented with various iterations of IBAFormer and observed that substituting the module within the first Transformer block and introducing intra-batch modules just before the decoder minimally affect the pre-trained model, as illustrated in Figure~\ref{fig:framework}.
Table~\ref{tab:mcba} and~\ref{tab:ecba} show the effectiveness of detailed components in IBAFormer with MIBA and EIBA, respectively. 
We explore the replacement of self-attention modules with intra-batch attention in two specific blocks (block1 and block2) of SegFormer.
Additionally, we investigate the impact of intra-batch attention before sending the features to the decoder.
The baseline model in the two tables represents the ``Baseline+RICA" from Table~\ref{tab:ablation_sum}. 
For MIBA in Table~\ref{tab:mcba}, replacing self-attention modules with intra-batch attention modules in block1, along with introducing four intra-batch attention modules before feature decoding, yields performance improvements. These architectural modifications lead to a performance boost of +0.9\% and +1.26\%, respectively.
While replacing block1 results in enhanced performance, replacing block2 shows degraded performance, {possibly because lower-level features are more sensitive to noise and bias hence intra-batch attention is more effective with them.}
We train intra-batch attention modules from scratch and utilize pre-trained self-attention module parameters from SegFormer, avoiding the need for retraining on ImageNet.
Our results highlight the effective capture of inter-sample information, contributing to a significant overall performance gain of +1.8\% in generalization compared to the base model.

Enhancing EIBA-based IBAFormer in Table~\ref{tab:ecba} yields similar findings. 
Substituting  EIBA modules in block1 and incorporating them before the decoder both enhance model transferability, resulting in performance gains of 1.1\% and 1.14\%, respectively. 
In summary, both MIBA-based and EIBA-based IBAFormers exhibit substantial enhancements over the baseline model, with improvements of 1.8\% and 2.25\%, respectively.




\begin{table}[t]
\caption{\textbf{Ablation studies} on (a) architecture design of MIBA-based IBAFormer, (b) architecture design of EIBA-based IBAFormer,  and (c) batch size comparison (G$\rightarrow$ C, B, M). The reported results are mIoU (\%).}
\label{tab:ablation_CBA}
\centering
\subtable[IBAFormer (MIBA)]{
\resizebox{\linewidth}{!}{
\begin{tabular}{ccc|ccc|c}
		\toprule[1pt]
		 block1&block2&decoder & {G$\rightarrow$ C }  &{G$\rightarrow$ B} &  {G$\rightarrow$ M} &{Average} \\   
         \toprule[1pt]
          $\times$ & $\times$ & $\times$ &  52.11& 48.75 & 56.76 & 52.54\\
         \hline 
          $\checked$ & $\times$&$\times$& 55.21 & 48.19 & 56.93 & 53.44\\
        \hline                 
        $\times$& $\checked$  &$\times$& 52.14 & 47.52 & 56.25 & 51.97 \\	
        \hline                 
        $\times$ & $\times$& $\checked$  & 53.95 & \textbf{49.46} & 58.00 & 53.80 \\	
        \hline  
        $\checked$  &$\times$ & $\checked$ & \textbf{55.29} & {49.16} & \textbf{58.58} & \textbf{54.34} \\				       
        \toprule[1pt]
\end{tabular}
}
\label{tab:mcba}
}
\subtable[IBAFormer (EIBA)]{
\resizebox{\linewidth}{!}{
\begin{tabular}{ccc|ccc|c}
		\toprule[1pt]
		 block1&block2&decoder & {G$\rightarrow$ C }  &{G$\rightarrow$ B} &  {G$\rightarrow$ M} &{Average} \\

         \toprule[1pt]
          $\times$ & $\times$ & $\times$ &  52.11& 48.75 & 56.76 & 52.54\\
         \hline 
          $\checked$ & $\times$&$\times$& 54.59 & 48.86 & 57.46 & 53.64\\
        \hline                 
        $\times$& $\checked$  &$\times$& 53.81 & 47.66 & 56.02 & 52.50 \\	
        \hline                 
        $\times$ & $\times$& $\checked$  & 54.88 & 49.45 & 56.7 & 53.68 \\	
         \hline                 
        $\checked$  &$\times$ & $\checked$ & \textbf{56.34}& \textbf{49.76} & \textbf{58.26} & \textbf{54.79} \\				       
        \toprule[1pt]
\end{tabular}
}
\label{tab:ecba}
}
\subtable[Batch size]{
\resizebox{\linewidth}{!}{
\begin{tabular}{c|c|ccc|c}
		\toprule[1pt]
		 IBA&Batch size & {G$\rightarrow$ C }  &{G$\rightarrow$ B} &  {G$\rightarrow$ M} &{Average} \\       
         \toprule[1pt]
           MIBA & 2 & 54.99 & \textbf{49.28} & 57.78 & 54.02\\
         \hline 
          MIBA & 4 & {55.29} & {49.16} & \textbf{58.58} & \textbf{54.34}\\
        \hline   
        MIBA & 8 & \textbf{55.88} & 49.1 & 57.03 & 54.00\\
        \hline 
           EIBA & 2 &  55.76 & 49.63 & 57.90 & 54.43\\
         \hline 
          EIBA & 4 & \textbf{56.34}& \textbf{49.76} & \textbf{58.26} & \textbf{54.79}\\
        \hline   
        EIBA & 8 & 55.7 & 49.39 & 57.88 & 54.32\\			       
        \toprule[1pt]
\end{tabular}
}
\label{tab:batch_size}
}
\end{table}

\subsubsection{Batch Size sensitivity analysis} 
To investigate the model's sensitivity to different batch sizes, we compare batch sizes of 2, 4, and 8 and observe their impact on the experimental results in Table~\ref{tab:batch_size}. 
The results indicate that varying batch sizes have a limited impact, with a relatively minor discrepancy observed, and batch size of 4 demonstrates slightly superior performance compared to batch sizes of 2 and 8.
To obtain satisfactory performance while making the most of GPU resources, we report the results using a batch size of 4 for all experiments.


\section{Conclusion}

The paper introduces intra-batch attention as a powerful mechanism for integrating cross-sample information, enriching  model input with valuable contextual insights.
Subsequently, IBAFormer is proposed to leverage intra-batch attention modules and improves the transferability of Transformers in DGSS.
We emphasize the significance of the intra-batch attention module, which conveniently enhances model transferability by providing an alternative to commonly used self-attention mechanisms in existing architectures.
However, the potential applicability of the intra-batch attention module in other vision tasks remains unexplored in this paper, and awaits investigation in future studies.

\bibliography{aaai24}

\begin{thebibliography}{52}
\providecommand{\natexlab}[1]{#1}

\bibitem[{Carion et~al.(2018)Carion, Massa, Synnaeve, Usunier, Kirillov, and Zagoruyko}]{nicola2018end}
Carion, N.; Massa, F.; Synnaeve, G.; Usunier, N.; Kirillov, A.; and Zagoruyko, S. 2018.
\newblock End-to-End Object Detection with Transformers.
\newblock In \emph{Proceedings of the European Conference on Computer Vision (ECCV)}, 464--479.

\bibitem[{Chen, Fan, and Panda(2021)}]{chen2021crossvit}
Chen, C.-F.; Fan, Q.; and Panda, R. 2021.
\newblock CrossViT: Cross-Attention Multi-Scale Vision Transformer for Image Classification.
\newblock \emph{arXiv preprint arXiv:2103.14899}.

\bibitem[{Cheng et~al.(2022)Cheng, Misra, Schwing, Kirillov, and Girdhar}]{cheng2022masked}
Cheng, B.; Misra, I.; Schwing, A.~G.; Kirillov, A.; and Girdhar, R. 2022.
\newblock Masked-attention mask transformer for universal image segmentation.
\newblock In \emph{Proceedings of the IEEE/CVF Conference on Computer Vision and Pattern Recognition (CVPR)}, 1290--1299.

\bibitem[{Choi et~al.(2021)Choi, Jung, Yun, Kim, Kim, and Choo}]{choi2021robustnet}
Choi, S.; Jung, S.; Yun, H.; Kim, J.~T.; Kim, S.; and Choo, J. 2021.
\newblock RobustNet: Improving Domain Generalization in Urban-Scene Segmentation via Instance Selective Whitening.
\newblock In \emph{Proceedings of the IEEE/CVF Conference on Computer Vision and Pattern Recognition (CVPR)}, 11580--11590.

\bibitem[{Cordts et~al.(2016)Cordts, Omran, Ramos, Rehfeld, Enzweiler, Benenson, Franke, Roth, and Schiele}]{cordts2016cityscapes}
Cordts, M.; Omran, M.; Ramos, S.; Rehfeld, T.; Enzweiler, M.; Benenson, R.; Franke, U.; Roth, S.; and Schiele, B. 2016.
\newblock The cityscapes dataset for semantic urban scene understanding.
\newblock In \emph{Proceedings of the IEEE/CVF Conference on Computer Vision and Pattern Recognition (CVPR)}, 3213--3223.

\bibitem[{Ding et~al.(2023)Ding, Xue, Xia, Schiele, and Dai}]{ding2023hgformer}
Ding, J.; Xue, N.; Xia, G.-S.; Schiele, B.; and Dai, D. 2023.
\newblock HGFormer: Hierarchical Grouping Transformer for Domain Generalized Semantic Segmentation.
\newblock In \emph{Proceedings of the IEEE/CVF Conference on Computer Vision and Pattern Recognition (CVPR)}, 15413--15423.

\bibitem[{Dosovitskiy et~al.(2021)Dosovitskiy, Beyer, Kolesnikov, Weissenborn, Zhai, Unterthiner, Dehghani, Minderer, Heigold, Gelly, Uszkoreit, and Houlsby}]{dosovitskiy2021an}
Dosovitskiy, A.; Beyer, L.; Kolesnikov, A.; Weissenborn, D.; Zhai, X.; Unterthiner, T.; Dehghani, M.; Minderer, M.; Heigold, G.; Gelly, S.; Uszkoreit, J.; and Houlsby, N. 2021.
\newblock An Image is Worth 16x16 Words: Transformers for Image Recognition at Scale.
\newblock In \emph{International Conference on Learning Representations (ICLR)}.

\bibitem[{Guo, Stutz, and Schiele(2023)}]{guo2023improving}
Guo, Y.; Stutz, D.; and Schiele, B. 2023.
\newblock Improving robustness of vision transformers by reducing sensitivity to patch corruptions.
\newblock In \emph{Proceedings of the IEEE/CVF Conference on Computer Vision and Pattern Recognition (CVPR)}, 4108--4118.

\bibitem[{Hoffman et~al.(2018)Hoffman, Tzeng, Park, Zhu, Isola, Saenko, Efros, and Darrell}]{hoffman2018cycada}
Hoffman, J.; Tzeng, E.; Park, T.; Zhu, J.-Y.; Isola, P.; Saenko, K.; Efros, A.; and Darrell, T. 2018.
\newblock Cycada: Cycle-consistent adversarial domain adaptation.
\newblock In \emph{International Conference on Machine Learning (ICML)}, 1989--1998. Pmlr.

\bibitem[{Hoyer, Dai, and Van~Gool(2022)}]{hoyer2022daformer}
Hoyer, L.; Dai, D.; and Van~Gool, L. 2022.
\newblock Daformer: Improving network architectures and training strategies for domain-adaptive semantic segmentation.
\newblock In \emph{Proceedings of the IEEE/CVF Conference on Computer Vision and Pattern Recognition (CVPR)}, 9924--9935.

\bibitem[{Huang et~al.(2021)Huang, Guan, Xiao, and Lu}]{huang2021fsdr}
Huang, J.; Guan, D.; Xiao, A.; and Lu, S. 2021.
\newblock Fsdr: Frequency space domain randomization for domain generalization.
\newblock In \emph{Proceedings of the IEEE/CVF Conference on Computer Vision and Pattern Recognition (CVPR)}, 6891--6902.

\bibitem[{Huang et~al.(2023)Huang, Chen, Li, Li, Li, Song, Yan, and Xiong}]{huang2023styleprojection}
Huang, W.; Chen, C.; Li, Y.; Li, J.; Li, C.; Song, F.; Yan, Y.; and Xiong, Z. 2023.
\newblock Style Projected Clustering for Domain Generalized Semantic Segmentation.
\newblock In \emph{Proceedings of the IEEE/CVF Conference on Computer Vision and Pattern Recognition (CVPR)}, 3061--3071.

\bibitem[{Huang and Belongie(2017)}]{huang2017arbitrary}
Huang, X.; and Belongie, S. 2017.
\newblock Arbitrary style transfer in real-time with adaptive instance normalization.
\newblock In \emph{Proceedings of the IEEE International Conference on Computer Vision (ICCV)}, 1501--1510.

\bibitem[{Kim et~al.(2022)Kim, Lee, Park, Min, and Sohn}]{kim2022pin}
Kim, J.; Lee, J.; Park, J.; Min, D.; and Sohn, K. 2022.
\newblock Pin the memory: Learning to generalize semantic segmentation.
\newblock In \emph{Proceedings of the IEEE/CVF Conference on Computer Vision and Pattern Recognition}, 4350--4360.

\bibitem[{Kim et~al.(2021)Kim, Son, Lan, Zeng, and Kwak}]{kim2021wedge}
Kim, N.; Son, T.; Lan, C.; Zeng, W.; and Kwak, S. 2021.
\newblock Wedge: web-image assisted domain generalization for semantic segmentation.
\newblock \emph{arXiv preprint arXiv:2109.14196}.

\bibitem[{Lee et~al.(2022)Lee, Seong, Lee, and Kim}]{lee2022wildnet}
Lee, S.; Seong, H.; Lee, S.; and Kim, E. 2022.
\newblock WildNet: Learning Domain Generalized Semantic Segmentation from the Wild.
\newblock In \emph{Proceedings of the IEEE/CVF Conference on Computer Vision and Pattern Recognition (CVPR)}, 9936--9946.

\bibitem[{Li et~al.(2023)Li, Chen, Li, Ding, and Zhao}]{li2023bvit}
Li, N.; Chen, Y.; Li, W.; Ding, Z.; and Zhao, D. 2023.
\newblock BViT: Broad Attention based Vision Transformer.
\newblock \emph{arXiv preprint arXiv:2202.06268}.

\bibitem[{Li, Shi, and Lin(2018)}]{Li_2018_CVPR}
Li, Y.; Shi, J.; and Lin, D. 2018.
\newblock Low-Latency Video Semantic Segmentation.
\newblock In \emph{Proceedings of the IEEE Conference on Computer Vision and Pattern Recognition (CVPR)}.

\bibitem[{Liu et~al.(2021)Liu, Lin, Cao, Hu, Wei, Zhang, Lin, and Guo}]{liu2021swin}
Liu, Z.; Lin, Y.; Cao, Y.; Hu, H.; Wei, Y.; Zhang, Z.; Lin, S.; and Guo, B. 2021.
\newblock Swin Transformer: Hierarchical Vision Transformer using Shifted Windows.
\newblock In \emph{Proceedings of the IEEE/CVF International Conference on Computer Vision (ICCV)}, 9992--10002.

\bibitem[{Loshchilov and Hutter(2017)}]{loshchilov2017decoupled}
Loshchilov, I.; and Hutter, F. 2017.
\newblock Decoupled weight decay regularization.
\newblock \emph{arXiv preprint arXiv:1711.05101}.

\bibitem[{Ma et~al.(2021)Ma, Lin, Wu, and Yu}]{ma2021coarse}
Ma, H.; Lin, X.; Wu, Z.; and Yu, Y. 2021.
\newblock Coarse-to-Fine Domain Adaptive Semantic Segmentation with Photometric Alignment and Category-Center Regularization.
\newblock In \emph{Proceedings of the IEEE/CVF Conference on Computer Vision and Pattern Recognition (CVPR)}, 4051--4060.

\bibitem[{Neuhold et~al.(2017)Neuhold, Ollmann, Rota~Bulo, and Kontschieder}]{neuhold2017mapillary}
Neuhold, G.; Ollmann, T.; Rota~Bulo, S.; and Kontschieder, P. 2017.
\newblock The mapillary vistas dataset for semantic understanding of street scenes.
\newblock In \emph{Proceedings of the IEEE/CVF International Conference on Computer Vision (ICCV)}, 4990--4999.

\bibitem[{Pan et~al.(2018)Pan, Luo, Shi, and Tang}]{pan2018two}
Pan, X.; Luo, P.; Shi, J.; and Tang, X. 2018.
\newblock Two at once: Enhancing learning and generalization capacities via ibn-net.
\newblock In \emph{Proceedings of the European Conference on Computer Vision (ECCV)}, 464--479.

\bibitem[{Paul and Chen(2022)}]{paul2022vision}
Paul, S.; and Chen, P.-Y. 2022.
\newblock Vision transformers are robust learners.
\newblock In \emph{Proceedings of the AAAI conference on Artificial Intelligence (AAAI)}, volume~36, 2071--2081.

\bibitem[{Peng et~al.(2022)Peng, Lei, Hayat, Guo, and Li}]{peng2022semantic}
Peng, D.; Lei, Y.; Hayat, M.; Guo, Y.; and Li, W. 2022.
\newblock Semantic-aware domain generalized segmentation.
\newblock In \emph{Proceedings of the IEEE/CVF Conference on Computer Vision and Pattern Recognition (CVPR)}, 2594--2605.

\bibitem[{Peng et~al.(2021)Peng, Lei, Liu, Zhang, and Liu}]{peng2021global}
Peng, D.; Lei, Y.; Liu, L.; Zhang, P.; and Liu, J. 2021.
\newblock Global and local texture randomization for synthetic-to-real semantic segmentation.
\newblock \emph{IEEE Transactions on Image Processing (TIP)}, 30: 6594--6608.

\bibitem[{Richter et~al.(2016)Richter, Vineet, Roth, and Koltun}]{richter2016playing}
Richter, S.~R.; Vineet, V.; Roth, S.; and Koltun, V. 2016.
\newblock Playing for data: Ground truth from computer games.
\newblock In \emph{Proceedings of the European Conference on Computer Vision (ECCV)}, 102--118. Springer.

\bibitem[{Ros et~al.(2016)Ros, Sellart, Materzynska, Vazquez, and Lopez}]{ros2016synthia}
Ros, G.; Sellart, L.; Materzynska, J.; Vazquez, D.; and Lopez, A.~M. 2016.
\newblock The synthia dataset: A large collection of synthetic images for semantic segmentation of urban scenes.
\newblock In \emph{Proceedings of the IEEE/CVF Conference on Computer Vision and Pattern Recognition (CVPR)}, 3234--3243.

\bibitem[{Sakaridis, Dai, and Van~Gool(2021)}]{sakaridis2021acdc}
Sakaridis, C.; Dai, D.; and Van~Gool, L. 2021.
\newblock ACDC: The adverse conditions dataset with correspondences for semantic driving scene understanding.
\newblock In \emph{Proceedings of the IEEE/CVF International Conference on Computer Vision (ICCV)}, 10765--10775.

\bibitem[{Seidenschwarz, Elezi, and Leal-Taix{\'e}(2021)}]{seidenschwarz2021learning}
Seidenschwarz, J.~D.; Elezi, I.; and Leal-Taix{\'e}, L. 2021.
\newblock Learning intra-batch connections for deep metric learning.
\newblock In \emph{International Conference on Machine Learning}, 9410--9421. PMLR.

\bibitem[{Sengupta, Budvytis, and Cipolla(2020)}]{STRAPS2020BMVC}
Sengupta, A.; Budvytis, I.; and Cipolla, R. 2020.
\newblock Synthetic Training for Accurate 3D Human Pose and Shape Estimation in the Wild.
\newblock In \emph{British Machine Vision Conference (BMVC)}.

\bibitem[{Sun et~al.(2023)Sun, Melnyk, Felsberg, and Tang}]{sun2023augment}
Sun, Q.; Melnyk, P.; Felsberg, M.; and Tang, Y. 2023.
\newblock Augment Features Beyond Color for Domain Generalized Segmentation.
\newblock \emph{arXiv preprint arXiv:2307.01703}.

\bibitem[{Tang et~al.(2020)Tang, Gao, Zhu, Zhang, Li, and Metaxas}]{tang2020selfnorm}
Tang, Z.; Gao, Y.; Zhu, Y.; Zhang, Z.; Li, M.; and Metaxas, D.~N. 2020.
\newblock Selfnorm and crossnorm for out-of-distribution robustness.

\bibitem[{Tjio et~al.(2022{\natexlab{a}})Tjio, Liu, Zhou, and Goh}]{tjio2022adversarial}
Tjio, G.; Liu, P.; Zhou, J.~T.; and Goh, R. S.~M. 2022{\natexlab{a}}.
\newblock Adversarial semantic hallucination for domain generalized semantic segmentation.
\newblock In \emph{Proceedings of the IEEE/CVF Winter Conference on Applications of Computer Vision (WACV)}, 318--327.

\bibitem[{Tjio et~al.(2022{\natexlab{b}})Tjio, Liu, Zhou, and Mong~Goh}]{tijio2022ash}
Tjio, G.; Liu, P.; Zhou, J.~T.; and Mong~Goh, R.~S. 2022{\natexlab{b}}.
\newblock Adversarial Semantic Hallucination for Domain Generalized Semantic Segmentation.
\newblock In \emph{Proceedings of the IEEE/CVF Winter Conference on Applications of Computer Vision (WACV)}, 3849--3858.

\bibitem[{Torralba and Efros(2011)}]{torralba2011unbiased}
Torralba, A.; and Efros, A.~A. 2011.
\newblock Unbiased look at dataset bias.
\newblock In \emph{CVPR 2011}, 1521--1528. IEEE.

\bibitem[{Wang et~al.(2021)Wang, Xie, Li, Fan, Song, Liang, Lu, Luo, and Shao}]{wang2021pyramid}
Wang, W.; Xie, E.; Li, X.; Fan, D.-P.; Song, K.; Liang, D.; Lu, T.; Luo, P.; and Shao, L. 2021.
\newblock Pyramid Vision Transformer: A Versatile Backbone for Dense Prediction without Convolutions.
\newblock In \emph{Proceedings of the IEEE/CVF International Conference on Computer Vision (ICCV)}, 548--558.

\bibitem[{Wang et~al.(2022)Wang, Yao, Chen, Lin, Cai, He, and Liu}]{wang2022crossformer}
Wang, W.; Yao, L.; Chen, L.; Lin, B.; Cai, D.; He, X.; and Liu, W. 2022.
\newblock CrossFormer: A Versatile Vision Transformer Hinging on Cross-scale Attention.
\newblock In \emph{International Conference on Learning Representations (ICLR)}.

\bibitem[{Wenzel et~al.(2022)Wenzel, Dittadi, Gehler, Simon-Gabriel, Horn, Zietlow, Kernert, Russell, Brox, Schiele et~al.}]{wenzel2022assaying}
Wenzel, F.; Dittadi, A.; Gehler, P.; Simon-Gabriel, C.-J.; Horn, M.; Zietlow, D.; Kernert, D.; Russell, C.; Brox, T.; Schiele, B.; et~al. 2022.
\newblock Assaying out-of-distribution generalization in transfer learning.
\newblock \emph{Advances in Neural Information Processing Systems (NeurIPS)}, 35: 7181--7198.

\bibitem[{Wu et~al.(2021)Wu, Xiao, Codella, Liu, Dai, Yuan, and Zhang}]{wu2021cvt}
Wu, H.; Xiao, B.; Codella, N.; Liu, M.; Dai, X.; Yuan, L.; and Zhang, L. 2021.
\newblock CvT: Introducing Convolutions to Vision Transformers.
\newblock \emph{arXiv preprint arXiv:2103.15808}.

\bibitem[{Wu et~al.(2022)Wu, Wu, Zhang, Ju, and Wang}]{wu2022siamdoge}
Wu, Z.; Wu, X.; Zhang, X.; Ju, L.; and Wang, S. 2022.
\newblock SiamDoGe: Domain Generalizable Semantic Segmentation Using Siamese Network.
\newblock In \emph{Proceedings of the European Conference on Computer Vision (ECCV)}, 603--620. Springer.

\bibitem[{Xie et~al.(2021)Xie, Wang, Yu, Anandkumar, Alvarez, and Luo}]{xie2021segformer}
Xie, E.; Wang, W.; Yu, Z.; Anandkumar, A.; Alvarez, J.~M.; and Luo, P. 2021.
\newblock SegFormer: Simple and efficient design for semantic segmentation with transformers.
\newblock \emph{Advances in Neural Information Processing Systems (NeurIPS)}, 34: 12077--12090.

\bibitem[{Xu et~al.(2022{\natexlab{a}})Xu, Zhang, Cai, Rezatofighi, and Tao}]{xu2022gmflow}
Xu, H.; Zhang, J.; Cai, J.; Rezatofighi, H.; and Tao, D. 2022{\natexlab{a}}.
\newblock Gmflow: Learning optical flow via global matching.
\newblock In \emph{Proceedings of the IEEE/CVF Conference on Computer Vision and Pattern Recognition (CVPR)}, 8121--8130.

\bibitem[{Xu et~al.(2022{\natexlab{b}})Xu, Zhang, Cai, Rezatofighi, Yu, Tao, and Geiger}]{xu2022unifying}
Xu, H.; Zhang, J.; Cai, J.; Rezatofighi, H.; Yu, F.; Tao, D.; and Geiger, A. 2022{\natexlab{b}}.
\newblock Unifying flow, stereo and depth estimation.
\newblock \emph{arXiv preprint arXiv:2211.05783}.

\bibitem[{Xu et~al.(2022{\natexlab{c}})Xu, Yao, Jiang, Jiang, Chu, Han, Zhang, Wang, and Tai}]{xu2022dirl}
Xu, Q.; Yao, L.; Jiang, Z.; Jiang, G.; Chu, W.; Han, W.; Zhang, W.; Wang, C.; and Tai, Y. 2022{\natexlab{c}}.
\newblock DIRL: Domain-invariant representation learning for generalizable semantic segmentation.
\newblock In \emph{Proceedings of the AAAI Conference on Artificial Intelligence (AAAI)}, volume~36, 2884--2892.

\bibitem[{Yu et~al.(2020)Yu, Chen, Wang, Xian, Chen, Liu, Madhavan, and Darrell}]{yu2020bdd100k}
Yu, F.; Chen, H.; Wang, X.; Xian, W.; Chen, Y.; Liu, F.; Madhavan, V.; and Darrell, T. 2020.
\newblock Bdd100k: A diverse driving dataset for heterogeneous multitask learning.
\newblock In \emph{Proceedings of the IEEE/CVF conference on Computer Vision and Pattern Recognition (CVPR)}, 2636--2645.

\bibitem[{Yuan et~al.(2021)Yuan, Chen, Wang, Yu, Shi, Jiang, Tay, Feng, and Yan}]{yuan2021tokenstotoken}
Yuan, L.; Chen, Y.; Wang, T.; Yu, W.; Shi, Y.; Jiang, Z.; Tay, F.~E.; Feng, J.; and Yan, S. 2021.
\newblock Tokens-to-Token ViT: Training Vision Transformers from Scratch on ImageNet.
\newblock \emph{arXiv preprint arXiv:2101.11986}.

\bibitem[{Yue et~al.(2019)Yue, Zhang, Zhao, Sangiovanni-Vincentelli, Keutzer, and Gong}]{yue2019domain}
Yue, X.; Zhang, Y.; Zhao, S.; Sangiovanni-Vincentelli, A.; Keutzer, K.; and Gong, B. 2019.
\newblock Domain randomization and pyramid consistency: Simulation-to-real generalization without accessing target domain data.
\newblock In \emph{Proceedings of the IEEE/CVF International Conference on Computer Vision (ICCV)}, 2100--2110.

\bibitem[{Zhao et~al.(2022{\natexlab{a}})Zhao, Zhong, Zhao, Sebe, and Lee}]{zhao2022styleextend}
Zhao, Y.; Zhong, Z.; Zhao, N.; Sebe, N.; and Lee, G.~H. 2022{\natexlab{a}}.
\newblock Style-Hallucinated Dual Consistency Learning: A Unified Framework for Visual Domain Generalization.
\newblock \emph{arXiv preprint arXiv:2212.09068}.

\bibitem[{Zhao et~al.(2022{\natexlab{b}})Zhao, Zhong, Zhao, Sebe, and Lee}]{zhao2022style}
Zhao, Y.; Zhong, Z.; Zhao, N.; Sebe, N.; and Lee, G.~H. 2022{\natexlab{b}}.
\newblock Style-hallucinated dual consistency learning for domain generalized semantic segmentation.
\newblock In \emph{Proceedings of the European Conference on Computer Vision (ECCV)}, 535--552. Springer.

\bibitem[{Zhong et~al.(2022)Zhong, Zhao, Lee, and Sebe}]{zhong2022adversarial}
Zhong, Z.; Zhao, Y.; Lee, G.~H.; and Sebe, N. 2022.
\newblock Adversarial style augmentation for domain generalized urban-scene segmentation.
\newblock \emph{Advances in Neural Information Processing Systems (NeurIPS)}, 35: 338--350.

\bibitem[{Zhou et~al.(2022)Zhou, Yu, Xie, Xiao, Anandkumar, Feng, and Alvarez}]{zhou2022understanding}
Zhou, D.; Yu, Z.; Xie, E.; Xiao, C.; Anandkumar, A.; Feng, J.; and Alvarez, J.~M. 2022.
\newblock Understanding the robustness in vision transformers.
\newblock In \emph{International Conference on Machine Learning (ICML)}, 27378--27394. PMLR.

\end{thebibliography}

\end{document}